\documentclass[conference]{IEEEtran}
\IEEEoverridecommandlockouts
\usepackage{cite}
\usepackage{amsmath,amssymb,amsfonts}
\usepackage{algorithmic}
\usepackage{graphicx}
\usepackage{textcomp}
\usepackage{xcolor}
\usepackage{todonotes}
\usepackage{hyperref}
\def\BibTeX{{\rm B\kern-.05em{\sc i\kern-.025em b}\kern-.08em
    T\kern-.1667em\lower.7ex\hbox{E}\kern-.125emX}}
\begin{document}

\title{Complex Vehicle Routing with Memory Augmented Neural Networks\\
\thanks{This work is being funded by the EU ECSEL Joint Undertaking under grant agreement no. 737459 (project Productive4.0).}
}

\author{\IEEEauthorblockN{1\textsuperscript{st} Marijn van Knippenberg}
\IEEEauthorblockA{\textit{Department of Mathematics and Computer Science} \\
\textit{Eindhoven University of Technology}\\
Eindhoven, Netherlands \\
m.s.v.knippenberg@tue.nl}
\and
\IEEEauthorblockN{2\textsuperscript{nd} Mike Holenderski}
\IEEEauthorblockA{\textit{Department of Mathematics and Computer Science} \\
\textit{Eindhoven University of Technology}\\
Eindhoven, Netherlands \\
m.holenderski@tue.nl}
\and
\IEEEauthorblockN{3\textsuperscript{rd} Vlado Menkovski}
\IEEEauthorblockA{\textit{Department of Mathematics and Computer Science} \\
\textit{Eindhoven University of Technology}\\
Eindhoven, Netherlands \\
v.menkovski@tue.nl}
}

\maketitle

\begin{abstract}
Complex real-life routing challenges can be modeled as variations of well-known combinatorial optimization problems. These routing problems have long been studied and are difficult to solve at scale. The particular setting may also make exact formulation difficult. Deep Learning offers an increasingly attractive alternative to traditional solutions, which mainly revolve around the use of various heuristics. Deep Learning may provide solutions which are less time-consuming and of higher quality at large scales, as it generally does not need to generate solutions in an iterative manner, and Deep Learning models have shown a surprising capacity for solving complex tasks in recent years. Here we consider a particular variation of the Capacitated Vehicle Routing (CVRP) problem and investigate the use of Deep Learning models with explicit memory components. Such memory components may help in gaining insight into the model's decisions as the memory and operations on it can be directly inspected at any time, and may assist in scaling the method to such a size that it becomes viable for industry settings.
\end{abstract}


\section{Introduction}
Combinatorial optimization problems are attractive targets because they translate to important real-life settings. In the area of logistics, packing and routing problems are often-considered with the goal of minimizing running costs. A typical example is that of a fleet of trucks being charged with delivering certain amounts of goods to a number of locations from a central depot. These problems are generally NP-hard, and solving them usually relies on some heuristic iterative method. Deep Learning is becoming an increasingly attractive method for solving these kinds of problems \cite{DBLP:journals/corr/abs-1811-06128,kong2018new}. One of the main advantages of Deep Learning is that once a model has been trained, generating routing solutions is much faster than using traditional methods. 

The routing problems can be ordered by the complexity of their task. The simplest is the Traveling Salesman Problem (TSP), which involves only one truck, and one constraint on the round-trip route (each point has to be visited exactly once). TSPs can be expanded by adding more trucks, becoming a Multiple Travelling Salesman Problem (MTSP) or Vehicle Routing Problem (VRP). Setting limitations on the capacity of the trucks turns it into a Capacitated Vehicle Routing Problem (CVRP). This problem then has numerous variations that add additional challenges. Examples of these include: time windows for trucks, time windows for pickup point, multiple depots instead of a single depot, compartmentalized trucks, trailers, and transshipment points. 

In this work, we consider a use case in the form of a CVRP with extensions where a fleet of trucks has to pick up cargo from a number of pickup points, and return this cargo to a single depot. This depot is also the point from which all trucks depart to start their route. Each truck has a limited carrying capacity, and each pickup point an amount of cargo waiting to be picked up, which can differ day by day. The goal is to find a solution which can generate efficient routing schedules on a daily basis. Preferably the solution is flexible enough to allow for future extensions to the problem definition, such as time windows for both the trucks and pickup points, and possibly compartmentalized trucks, transshipment points, and multiple depots. Another possible extension is more accurate travelling costs i.e., truck operating costs being a function of its load weight. 

A problem instance is considered to be an undirected, weighted, complete graph that models the points that the trucks need to visit on their routes. The cost of travelling from one point to another is condensed into a single value which is stored in the edge between those two points. Graph nodes may have different attributes, such as the amount of cargo to be picked up, and the time window in which it is reachable. The same applies to the trucks in the fleet (capacity, and operating time window, for example). In the case of VRP-like problem instances, a fixed node (ID $0$) is considered to be the depot, which is where trucks start and end their route, and which acts as a sink for cargo they carry.

Before we tackle CVRP problems, we start by solving the more simpler problems, treating them as simplified versions of CVRPs. This guided approach gives us fixed evaluation points, and will reduce training times through the application of Transfer Learning.

\begin{figure*}[ht]
    \centering
    \includegraphics[width=14cm]{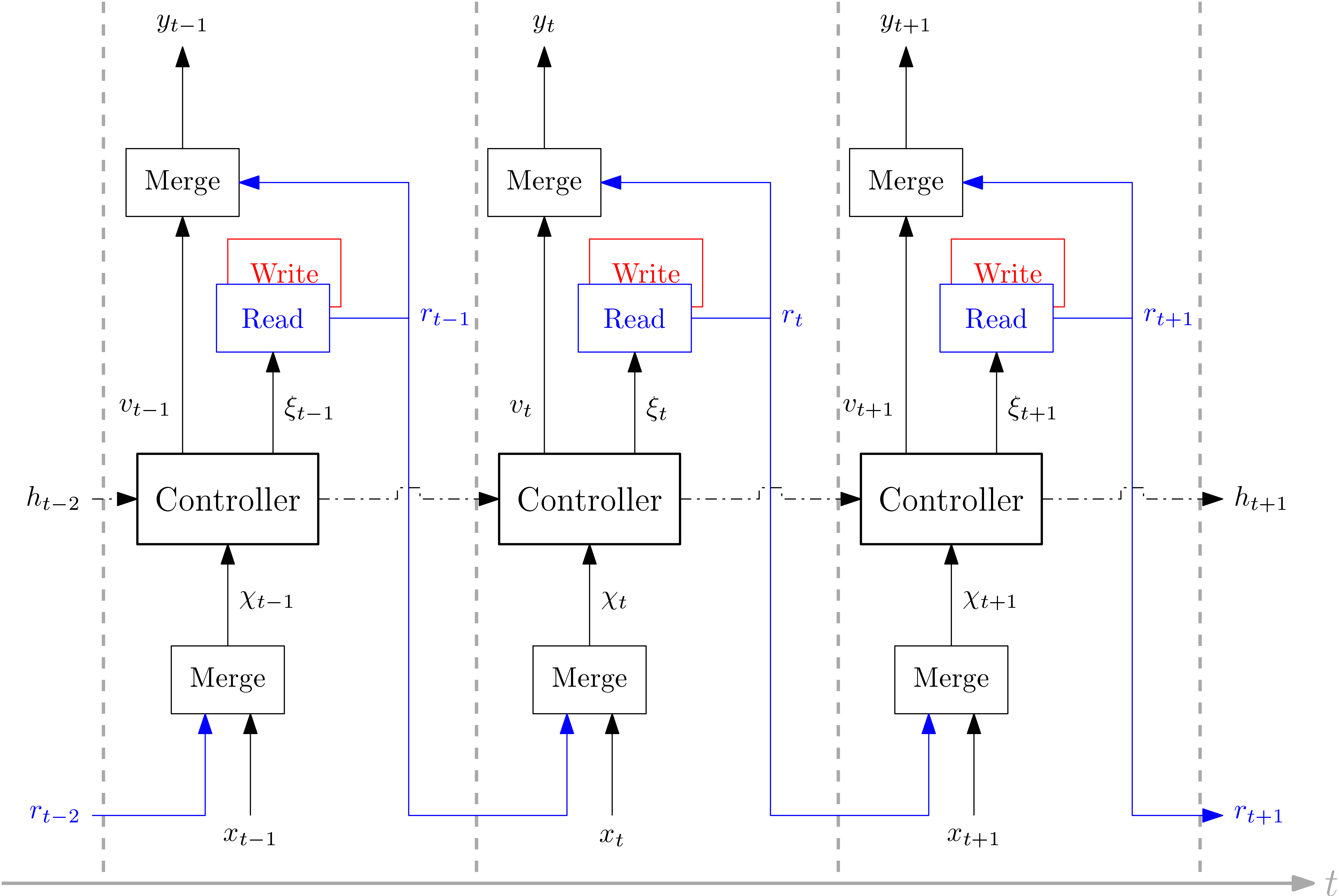}
    \caption{DNC workflow: The controller receives what was read from memory in the previous time step ($r$) in addition to the current input ($x$). It produces both output ($v$) and instructions for memory operations ($\xi$). After executing all memory operations, read results are both passed to the final output, and to the next time step.}
    \label{fig:dnc_flow}
\end{figure*}

\section{Background}

The idea to employ a Neural Network to solve NP-hard optimization problems came to the fore with the use of Hopfield Networks to solve TSP problems over two decades ago \cite{DBLP:journals/informs/Smith99}. Another major branch of research around this time focused on Self-Organising Feature Maps. 

While this provides a simple and compact format for problem instances, do note that a complete graph has a completely homogeneous structure. Therefore many graph learning methods which rely on graph structure will not work in this case. These kinds of methods are often applied to social network graphs, where each node is not connected to all other nodes \cite{DBLP:journals/corr/abs-1810-00826}. 

A number of previous attempts have been based on Reinforcement Learning \cite{DBLP:conf/nips/NazariOST18,Kool2018AttentionLT,DBLP:conf/nips/KhalilDZDS17}, but these do not scale effectively to problems with hundreds or even thousands of nodes, or are applied to only simple versions of the routing problem. Another challenge, which is caused by the application of Neural Networks, is the difficulty of working with sets as input, and finding solutions which are invariant to the order in the set is provided \cite{DBLP:journals/corr/abs-1803-09621,DBLP:journals/corr/abs-1802-04948}. These works often target TSP as an example problem, but do not try to solve it at large scale or in a more complex variation. Hybrid approaches have also been proposed, where a Deep Neural Network learns to generate a good heuristic, which is then used to solve the actual combinatorial optimization problem using a traditional, iterative method \cite{DBLP:journals/corr/abs-1903-03332,DBLP:journals/corr/abs-1709-09972,DBLP:conf/cpaior/DeudonCLAR18}.

\section{Method}

\begin{figure}[ht]
    \centering
    \includegraphics[width=0.3\textwidth]{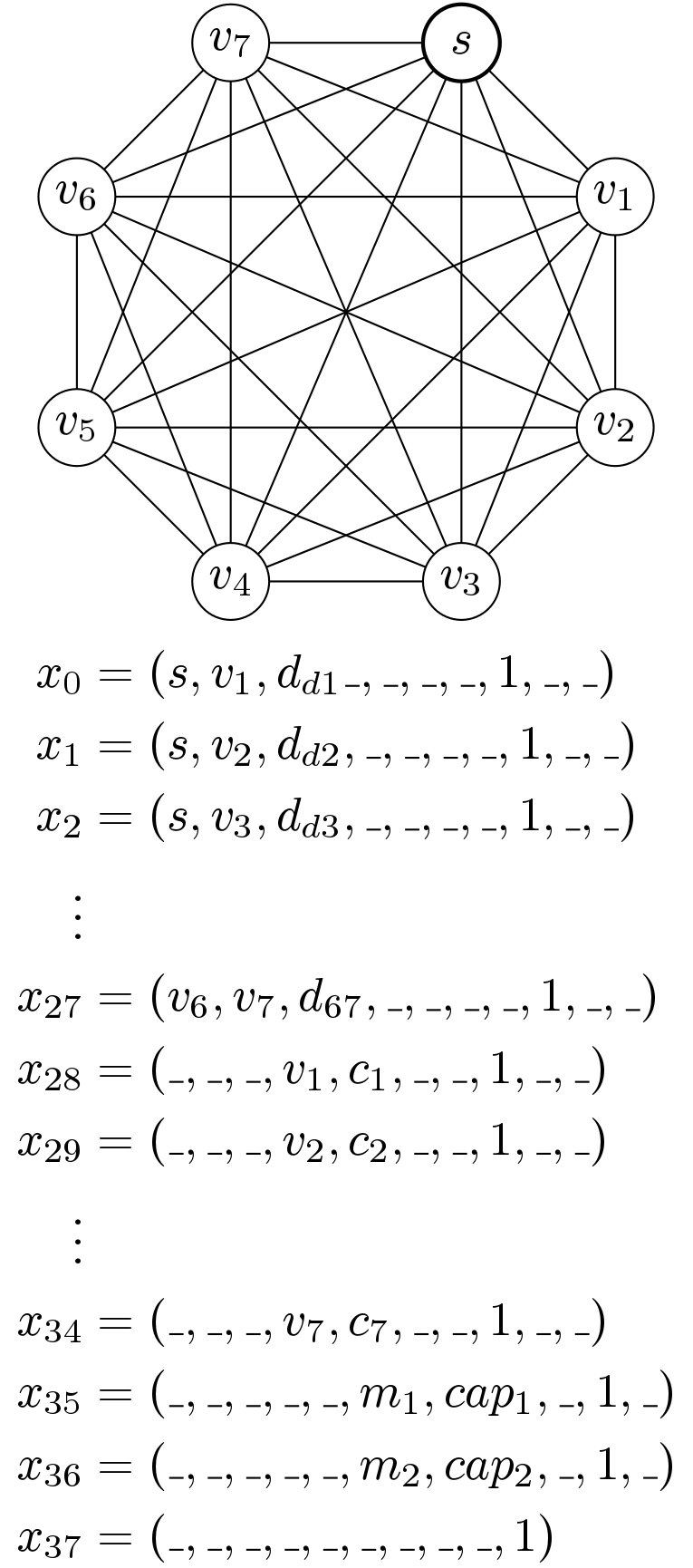}
    \caption{DNC input example: Example of input to a DNC for solving a CVRP on the graph above. During the first 28 time steps, the graphs itself is provided as a list of weighted edges. Next are the pickup amounts at each node. Then the task is provided as a list of trucks with their capacities (in this case 2 trucks). Finally, the DNC is instructed to start producing a solution. The last three values of each input are flags that indicate the type of input. The underscores indicate an empty value, which is a reserved number that indicates that the field is not filled (usually 0).}
    \label{fig:dnc_input}
\end{figure}

The approach in this work is based on the concept of Deep Learning models augmented with explicit memory modules, called Differentiable Neural Computers (DNCs) \cite{DBLP:journals/nature/GravesWRHDGCGRA16,DBLP:conf/nips/RaeHDHSWGL16}. The memory in this model is separate from the Neural Network, relieving the network of the responsibility of storing information regarding the current problem instance. The Neural Network is responsible for processing new input, generating operations for the memory, and generating the final output. The memory is integrated into the model such that during training back-propagation moves through both the network and the memory. During each time step, the network receives a part of the problem instance, such as a graph edge as well as data that was read from the memory in the previous time step. Combined, this data is propagated through the network and produces an output which is split into two parts. One of these parts contains the instructions for the memory in this time step. Based on these instructions, some memory is read, and some memory is written. The read results are then combined with the other output component to provide the final output of this time step. For a schematic overview of this process, see Figure \ref{fig:dnc_flow}.
During inference, the problem data is initially fed into the network, which will write it into the memory. Using a flag in the input stream, the input then switches to feeding in the task, which is also stored in memory. With another flag, the network is then signaled to start producing a solution. Additionally, the network may be given some time steps to process the input before requesting the solution. A simple example is provided in Figure \ref{fig:dnc_input}.

By being relieved of storing the problem setting in the network and having robust storage that can be accessed in a structured manner, this model has been shown to be able to handle complex input and solve complex problems. It is also more resistant to the influence of input order e.g., the order in which graph edges are fed into the network, than models such as pointer networks \cite{DBLP:conf/nips/VinyalsFJ15,DBLP:journals/corr/VinyalsBK15}, since initially such data is directly moved into the memory and can later be retrieved. In Neural Networks without explicit memory, this information is written directly into the weights in the network, being constantly overwritten as data is fed into the network. In DNCs memory usage is tracked and updated at each time step, which prevents the unintentional overwriting of information. Note also that this approach places no requirements on the input graph; it can be both non-Euclidean and asymmetric. It does however come with some disadvantages. Implementation is significantly more complex with the addition of the memory component and the mechanics for interacting with it. It also adds a number of parameters to the model that need to be carefully considered on a case by case basis. These include memory size, word size (size of a single memory unit), the number of read operations per time step, and the number of write operations per time step. Memory elements have to be large enough to uniquely encode all input elements (e.g., uniquely encode all edges in a graph, including any and all attributes). The memory itself has to contain enough units to contain both the current graph, and the intermediate solution state while the full solution is being generated.

\section{Experiments and Results}

Model parameters were taken directly from the results of the original DNC, including controller size, memory dimensions, and optimizer parameters. 
Given an undirected, weighted graph $G=(V, E)$, a vector of cargo amounts $cost$ and a vector of truck capacities $cap$, input for each problem instance is generated as a sequence of samples with the following structure

$$x = (v_i, v_j, d, v_k, cost_k, m_l, cap_l, data, task, solve)$$

, where $v_i$ and $v_j$ are node IDs, $d$ is the distance between these two nodes, $v_k$ is a node ID, $cost_k$ is the amount of cargo to be picked up at that node, $m_l$ is a truck ID, and $cap_l$ is the capacity of that truck. The 0-1 flags $data$, $task$, and $solve$ indicate the stage that the task is in. These fields are not all filled for each sample. Distance information, cargo amounts, and truck capacity are provided separately in a sequential manner. Other fields at these times are filled with a reserved value that indicates an empty field (indicated hereafter by an underscore). Providing an edge to the model may for example look like

$$x = (v_i, v_j, d, \_, \_, \_, \_, 1, \_, \_)$$

, while passing cargo information looks like

$$x = (\_, \_, \_, v_k, cost_k, \_, \_, 1, \_, \_)$$

By far the most input samples contain edge information, as their number scales quadratically in the number of nodes in the graph. Each additional pickup point and trucks require a single extra input sample for the cargo and capacity information. A solution can then be requested by setting the appropriate flag

$$x = (\_, \_, \_, \_, \_, \_, \_, \_, \_, 1)$$

This input is provided multiple times, the number of times depending on the expected size of the output (the number of edges that need to be traversed by the trucks in total). At each time step, the model will emit emit a single edge for a single truck. Together, these edges compose the route(s) of the truck(s).

For the Shortest Path Problem, the extra samples are generated, which contain the starting and end node of the rouse, respectively. Note that for the non-capacitated problems, the capacity of each truck is simply set to the sum of all cargo that needs to be picked up. Furthermore, for the problems that do not consider a fleet of trucks, it suffices to pass a single of such a truck as input. The pickup point that contains the depot has a fixed ID across problem instances and it is assumed that trucks always start from this depot. In the case of a Shortest Path Problem, the input is generated such that the starting node has the same ID as the depot has for the other problems. As routes are always circular, this format can thus represent TSPs, MTSPs, VRPs, and CVRPs.

Another limitation of DNCs is that particular approach to training called Curriculum Learning is required to efficiently train DNCs. Training from the start on large-scale instances is extremely inefficient, regardless of whether the size of the memory, as the model is still dependent on the controller generating proper instructions for the memory. It is much more efficient to instead start with small problem instances and then gradually increase the size and complexity, until the desired problem setting is reached. Generally, the DNC is trained on a certain problem size and evaluated at intervals on a hold-out set of problem instances of the same size. If the DNC performance on this set exceeds a preset threshold, it is advanced to the next ``lesson'', and repeats the process with slightly larger problem instances. Deciding the curriculum schedule adds yet more parameters to the network which influence the training cost. Small intervals between lesson size and/or complexity means unnecessarily many lessons, increasing training time. Making the step size between lessons too large on the other hand, means each individual lessons becomes harder, which may also increase training time. Clearly, there is a trade-off here that may be optimized for an ideal lesson ``pace''. Note that the ideal pace is dependant on the particular tasks that the model is training for, as well as the parameters chosen for the model. In the case of the described use case, Curriculum Learning is combined with a Transfer Learning approach, whereby a network trained on one task, is then further trained to perform another, similar task in order to decrease training time. Here, a curriculum is devised whereby not only the size of the problem instance increases, but also the complexity of the task in term of goals. Instead of immediately training in a complex CVRP setting, training may be more efficient by first training on simpler, but related problems, such as TSP, MTSP, and vanilla CVRP instances. Additionally, for production purposes, a model for solving vanilla CVRP problems may be used as a starting point for training models that are specialized in different CVRP variations.

For our current experiments, we follow the general approach of the original DNCs. Lessons are structured to steadily increase in complexity with some overlap between lessons. Graph size is randomized within bounds to achieve this overlap. Once $80\%$ of the problems in a hold-out set can be successfully solved, the lesson is finished and training moves on to the next lesson. At this time, the current accuracy for the current problem type is also recorded by checking how many problems it can solve of the hold-out set of problems of the last lesson of the current problem type. 

Training and testing instances can be generated synthetically. At its simplest (for Shortest Path Problems (SPP) and TSP problems) a sample is a symmetric distance matrix. For the Vehicle Routing class of problems, a fleet of trucks needs to be additionally generated. In general, the challenge here is to generate samples that resemble the topography of real-life infrastructure. Mainly, this means that the pickup points occur in certain patterns; clustered in the case of towns/cities and spread more evenly in the case of farms, for example. Truck fleets are mostly homogeneous in their composition. The industry partner for this project has provided some real samples of tasks on which the synthetic data can be sampled. As with the synthetic data, these consist of a symmetric distance matrix of all the pickup points and the depot, and a structured description of the attributes of the pickup points and the trucks in the fleet. These examples can be considered relatively stable. The amount of cargo at each pickup point may fluctuate day by day, but the pool of customers and fleet of trucks change only rarely. Given the relative simplicity of the data that has to be generated, a generating model was not considered here. Should the problem instances evolve to such a complexity that synthetic data generated at this level of simplicity starts impacting performance, a model that learns to generate instances may be reconsidered.

On small problem instances, the true labels against which the training model can be compared for the sake of optimization can be calculated exactly based on the generated graphs. For the Shortest Path Problems,  Dijkstra's algorithm \cite{dijkstra1959note} was used to compute the exact solutions. For the TSP instances up to 30 nodes, a straight-forward application of the Held-Karp algorithm \cite{held1962dynamic} was utilized to compute exact solutions. As the problem size grows during the curriculum it becomes unfeasible to compute exact solutions for thousands of instances. Here we must turn to heuristic-based solvers with time limits. The following are used

\begin{itemize}
    \item Concorde: A solver for TSP-like problems considered to be (one of) the fastest solvers currently in existence\footnote{\url{http://www.math.uwaterloo.ca/tsp/concorde/}}.
    \item LKH-3: A solver which supports many TSP and CVRP variations\footnote{\url{http://akira.ruc.dk/~keld/research/LKH-3/}} \cite{helsgaun2017extension}.
    \item Industry solution: The industry partner for this project currently has an in-house solver that is in production. This is the final benchmark against which to test the overall method. Though this solver only operates on CVRP problems, other problem types can be rewritten into CVRP problems.
\end{itemize}

Should these, too, become unfeasible for generating true labels, Reinforcement Learning may be applied to train the model without the need of labels. In this case, the model is directly trained on the total cost of the generated routing schedule, with the total cost acting as penalty. In a Reinforcement Learning setting, the state space $s$ is composed of the graph and the current location of the trucks in it, as well as the various attributes of the pickup points and the trucks. The action space $a$ is the choice of destination for each of the trucks. Choosing a destination is achieved by assigning a probability to each of the edges of a node at which a truck is located, and choosing the most likely edge. This will be the focus of future work.

The curriculum starts with the Traveling Salesman problem, as this is the least constrained of the problems. Training then moves to a multi-vehicle setting (VRP), followed by the addition of cargo demands and constraints (CVRP), and finally the addition of various other constraints. Initially, the only parameter besides the problem type for each lesson is the problem size in terms of graph size. Later problems introduce additional lesson parameters, such as truck fleet size, truck capacities, and pickup cargo amounts. 

\begin{table}[ht]
\centering
\caption{Example curriculum for routing problems. Node numbers are the bounds for the number of nodes for problem instances. Sizes are sampled uniformly at random. Test is the accuracy at the end of the lesson on the problem size of the last lesson of that particular problem type (lessons 4, 7, and 10, respectively). Note that different lessons take different amounts of time to reach their passing threshold.}
\begin{tabular}{rrrrr}
\hline
Lesson  & Problem & Nodes & Trucks & Test (\%) \\
\hline
1 & TSP & $(5, 10)$ && 24.6 \\
2 & TSP & $(5, 20)$ && 58.9 \\
3 & TSP & $(10, 20)$ && 84.0 \\
4 & TSP & $(10, 25)$ && 87.2 \\
5 & VRP & $(10, 20)$ && 19.2 \\
6 & VRP & $(10, 25)$ && 50.4 \\
7 & VRP & $(20, 30)$ && 74.9 \\
8 & CVRP & $(10, 25)$ & 2 & 12.4 \\
9 & CVRP & $(20, 30)$ & 3 & 58.4 \\
10 & CVRP & $(20, 30)$ & 4 & 72.8\\
\hline
\end{tabular}
\label{tab:curriculum}
\end{table}

Given enough training time, the model appears to be able to solve increasingly difficult instances of the routing problems, but it is the training times that are an issue. Even for the very modest problem sizes listed in above, the time required to train for a single lesson takes hours (for the TSP instances), if not days (for the VRP and CVRP instances), even on modern hardware (NVIDIA Tesla V100). Furthermore, subsequent lessons take increasingly long to complete unless the pass rate (80\% by default) is relaxed. Do note though that any application of a curriculum, no matter how rudimentary, decreases the total training time. Training only on final-size problems is incredibly inefficient, with total training times being potentially orders of magnitude larger.

\section{Conclusion}

A method for solving a family of common routing problems was presented and its performance was measured in a limited setting. Although it was shown to be able to solve these types of problems, even in the limited settings there are serious problems regarding training time. As new models and new techniques to tackle different data types are constantly surfacing, Deep Learning remains an attractive approach for the solving of routing problems, but memory-augmented networks may not be the most promising direction. As these problems are traditionally not targeted by the Deep Learning community, there is a wide berth of methods and techniques that remain to be evaluated in the context of solving these problems. In particular, various (Multi-Agent) Deef Reinforcement Learning (DRL) approaches show potential.

\section{Future Considerations}

Besides the focus on training times, there are a number of other directions of research that may be of interest. In this section, a number of future directions of work that are of particular interest are highlighted and described in short.

\subsection{Real-Time (Re-)Scheduling}
Traffic situations change constantly, affecting travel costs between nodes. Changes can be predictable (planned construction work), semi-predictable (extreme weather), or unpredictable (traffic accidents). Being able to quickly update a routing schedule based on incoming data is attractive in this use case as it can further help decrease operation costs by optimizing routing on the fly.

\subsection{Advanced Routing Possibilities}
In some settings, additional capabilities may be introduced for the trucks or the pickup points, or both. A good example is the concept of a transshipment point. Some pickup point may allow visiting trucks to temporarily store a trailer at their location for later retrieval by the same, or possibly a different, truck. Options like this may come with their own limitations, such as time constraints for storage and retrieval, or the number of trailers that can be stored at a location. Supporting such advanced options would further help in optimizing running costs.

\subsection{Read-Only Memory for Production}
In production, it can be reasonably assumed that there will be cases where a particular instance of the model is only used to generate routing schedules for a single problem, optionally with minor changes. This would be the case if a user only has one business operation that makes use of the model (a fixed pool of customers and a fixed truck fleet, for example), or if multiple operations each use their own instance of the model (specialized per operation). In such a case, it may not be necessary to feed the graph into the model each time during inference. Instead, the graph could be stored in a section of the model memory which is marked as read-only. This would prevent the model from changing or overwriting the graph data, while still having part of the memory available to work with for the intermediate solution. As the graph input constitutes the major part of the input, this could significantly reduce inference time.

\subsection{Advanced Lesson Scheduling}
Currently, lesson schedules are pre-made by hand and fixed during training. A possibly more effective approach would be to generate lessons ``on the fly'' (during training). Based on performance in lessons so far (looking back one or possibly more lessons), the step in complexity for the next lesson can be sized in an adaptive manner, in an effort to find the most optimal lesson schedule. Users would then only be required to provide the format of the first lesson, the criteria for the final lesson, and optionally bounds for the steps between lessons.

\subsection{Alternative Agent Modeling}

An attractive alternative to a single-agent setting is to spread the workload among multiple equal agents \cite{DBLP:journals/corr/abs-1812-11794}. In the particular use case highlighted in this work, each learning agent would represent a single truck, instead of one learning agent being responsible for the routing of all trucks in a scenario. This relieves the single agent of having to perform context-switching between trucks (in the case of a time-ordered output format), and reduces the individual agent size. Using smaller Neural Networks could furthermore greatly decrease training times. 

The major challenge in this approach lies in information sharing and synchronization. The agents need to be aware, to a certain extent, of each other's positions, capabilities, and intentions in order to generate qualitative solutions. Previous work has tackled this challenge among others by adding contextual information to Reinforcement Learning methods with some success \cite{DBLP:conf/kdd/LinZXZ18}. The use of DNCs however, adds another interesting approach to information sharing among agents. Because a DNC has an explicit memory component, this component could possibly be shared between all agents in a setting, providing the same context to all agents. Like processes with a shared memory, the agents would have access to a global context of all agents without the need for explicit message passing between individual agents.

\bibliographystyle{IEEEtran}
\bibliography{codl}

\end{document}